%% file: main.tex
\documentclass[11pt,a4paper]{article}
\usepackage[nohyperref]{template/emnlp-ijcnlp-2019}
\usepackage{times}
\usepackage{latexsym}
\usepackage{graphicx}
\usepackage{multirow}
\usepackage{soul}
\usepackage{subcaption}

\usepackage{url}

\aclfinalcopy 

\title{HARE: a Flexible Highlighting Annotator for Ranking and Exploration}

\author{
    Denis Newman-Griffis$^{\dagger,\ddagger}$ \and Eric Fosler-Lussier$^\dagger$\\
    $^\dagger$Dept.\ of Computer Science and Engineering, The Ohio State University, Columbus, OH\\
    $^\ddagger$Rehabilitation Medicine Dept., Clinical Center, National Institutes of Health, Bethesda, MD\\
  {\tt \{newman-griffis.1, fosler-lussier.1\}@osu.edu}
}

\date{}

\begin{document}
\maketitle
\begin{abstract}
    \input{abstract}
\end{abstract}

\input{sections/1-introduction}
\input{sections/2-background}
\input{sections/3-system-description}

\input{sections/4-results}

\input{sections/5-discussion}
\input{sections/6-conclusions}

\bibliography{references-cleaned}
\bibliographystyle{template/acl_natbib}

\clearpage
\appendix
\input{sections/appendix}

\end{document}

%% file: abstract.tex
Exploration and analysis of potential data sources is a significant challenge
in the application of NLP techniques to novel information domains. We describe
HARE, a system for highlighting relevant information in document collections
to support ranking and triage, which provides tools for post-processing and
qualitative analysis for model development and tuning. We apply HARE to the
use case of narrative descriptions of mobility information in clinical data,
and demonstrate its utility in comparing candidate embedding features. We
provide a web-based interface for annotation visualization and document
ranking, with a modular backend to support interoperability with existing
annotation tools.

%% file: sections/1-introduction.tex
\section{Introduction}

As natural language processing techniques become useful for an increasing
number of new information domains, it is not always clear how best to identify
information of interest, or to evaluate the output of automatic annotation
tools. This can be especially challenging when target data in the form of
long strings or narratives of complex structure, e.g., in financial data
\cite{Fisher2016} or clinical data \cite{Rosenbloom2011}.

We introduce HARE, a \ul{H}ighlighting \ul{A}nnotator for \ul{R}anking and
\ul{E}xploration. HARE includes two main components: a workflow for supervised
training of automated token-wise relevancy taggers, and a web-based
interface for visualizing and analyzing automated tagging output.
It is intended to serve two main purposes: (1) triage of documents
when analyzing new corpora for the presence of relevant information, and
(2) interactive analysis, post-processing, and comparison of output from
different annotation systems.

In this paper, we demonstrate an application of HARE to information about
individuals' mobility status, an important aspect of functioning concerned
with changing body position or location. This is a relatively new type of
health-related narrative information with largely uncharacterized linguistic
structure, and high relevance to overall health outcomes and work
disability programs. In
experiments on a corpus of 400 clinical records, we show that
with minimal tuning,
our tagger
is able to produce a high-quality ranking of documents based on their relevance
to mobility, and to capture mobility-likely document segments with high
fidelity. We further demonstrate the use of post-processing and qualitative
analytic components of our system to compare the impact of different feature
sets and tune processing settings to improve relevance tagging quality.

%% file: sections/2-background.tex
\section{Related work}

Corpus annotation tools are plentiful in NLP research: brat
\cite{Stenetorp2012} and Knowtator \cite{Ogren2006} being two heavily used
examples among many. However, the primary purpose of these tools is to
streamline \textit{manual} annotation by experts, and to support review
and revision of manual annotations. Some tools,
including brat, support automated pre-annotation,
but analysis of these annotations and corpus exploration is not commonly
included. Other tools, such as SciKnowMine,\footnote{
    \url{https://www.isi.edu/projects/sciknowmine/overview}
} use automated techniques for triage, but for routing to experts
for curation rather than ranking and model analysis. Document ranking
and search engines such as Apache Lucene,\footnote{
    \url{https://lucene.apache.org/}
} by contrast, can be overly fully-featured for early-stage
analysis of new datasets, and do not directly offer tools for annotation
and post-processing.

Early efforts towards extracting mobility information have
illustrated that it is often syntactically and semantically
complex, and difficult to extract reliably
\cite{Newman-Griffis2018a,Newman-Griffis2019}. Some characterization of
mobility-related terms has been performed as part of larger work on functioning
\cite{Skube2018}, but a lack of standardized terminologies limits the
utility of vocabulary-driven clinical NLP tools such as CLAMP \cite{Soysal2018}
or cTAKES \cite{Savova2010}. Thus, it forms a useful test case for HARE.

%% file: sections/3-system-description.tex
\input{tables/tbl-btris}

\section{System Description}

Our system has three stages for analyzing document sets, illustrated
in Figure~\ref{fig:workflow}. First, data
annotated by experts for token relevance can be used to train relevance
tagging models, and trained models can be applied to produce relevance scores
on new documents (Section~\ref{ssec:relevance-tagging-workflow}). Second,
we provide configurable post-processing tools for cleaning and smoothing
relevance scores (Section~\ref{sec:post-processing}). Finally, our system
includes interfaces for reviewing detailed relevance output, ranking documents
by their relevance to the target criterion, and analyzing
qualitative outcomes of relevance scoring output
(Sections~\ref{ssec:annotation-viewer}-\ref{ssec:qualitative-analysis});
all of these interfaces
allow interactive re-configuration of post-processing settings and switching
between output relevance scores from different models for comparison.

For our experiments on mobility information, we
use an extended version of the dataset described by \citet{Thieu2017}, which
consists of 400 English-language Physical
Therapy initial assessment and reassessment notes from the Rehabilitation
Medicine Department of the NIH Clinical Center. These text documents have been
annotated at the token level for descriptions and assessments of patient
mobility status. Further information on this dataset is given in
Table~\ref{tbl:btris}. We use ten-fold cross validation for our experiments,
splitting into folds at the document level.

\subsection{Relevance tagging workflow}
\label{ssec:relevance-tagging-workflow}

All hyperparameters discussed in this section were tuned on held-out
development data in cross-validation experiments. We report the best
settings here, and provide full comparison of hyperparameter settings
in Appendix~\ref{app:hyperparameters}.

\input{figures/fig-workflow}

\subsubsection{Preprocessing}

Different domains exhibit different
patterns in token and sentence structure that affect preprocessing.
In clinical text, tokenization is not a consensus issue, and a variety of
different tokenizers are used regularly \cite{Savova2010,Soysal2018}. As
mobility information is relatively unexplored, we relied on
general-purpose
tokenization with spaCy \cite{spaCy} as our default tokenizer, and WordPiece
\cite{Wu2016} for experiments using BERT. We did not apply sentence segmentation,
as clinical toolkits often produced short segments that
interrupted mobility information in our experiments.

\subsubsection{Feature extraction}

Our system supports feature extraction for individual tokens in input documents
using both static and contextualized word embeddings.

\textbf{Static embeddings} Using static (i.e., non-contextualized) embeddings,
we calculate input features for each token as the mean embedding of the token and
10 words on each side (truncated at sentence/line breaks). We used FastText
\cite{Bojanowski2017} embeddings trained on
a 10-year collection of physical and occupational therapy records from the NIH
Clinical Center.

\textbf{ELMo} \cite{Peters2018} ELMo features are calculated for each token by
taking the hidden states of the two bLSTM layers and the token layer,
multiplying each vector by learned weights,
and summing to produce a final embedding. Combination weights are trained
jointly with the token annotation model.
We used a 1024-dimensional ELMo model pretrained on PubMed data\footnote{
    \url{https://allennlp.org/elmo}
} for our mobility experiments. 

\textbf{BERT} \cite{Devlin2019} For BERT features, we take the hidden states of
the final $k$ layers of the model; as with ELMo embeddings, these outputs are
then multiplied by a learned weight vector, and the weighted layers are summed
to create the final embedding vectors.\footnote{
    Note that as BERT is constrained to use WordPiece tokenization, it may use
    slightly longer token sequences than the other methods.
} We used the 768-dimensional clinicalBERT \cite{Alsentzer2019} model\footnote{
    \url{https://github.com/EmilyAlsentzer/clinicalBERT}
} in our experiments, extracting features from the last 3 layers.

\input{figures/fig-thresholding}

\subsubsection{Automated token-level annotation}

We model the annotation process of assigning a relevance score for
each token using a feed-forward deep neural network that takes embedding
features as input and produces a binomial softmax distribution as output.
For mobility information, we used a DNN with three 300-dimensional hidden
layers, relu activation, and 60\% dropout. 

As shown in Table~\ref{tbl:btris}, our mobility dataset is considerably
imbalanced between relevant and irrelevant tokens. To adjust for this balance,
for each epoch of training, we used all of the relevant tokens
in the training documents, and sampled irrelevant tokens at a 75\% ratio
to produce a more balanced training set; negative points were
re-sampled at each epoch. As token predictions are conditionally independent
of one another given the embedding features, we did not maintain any sequence
in the samples drawn. Relevant samples were weighted at a ratio of 2:1 during
training.

After each epoch, we evaluate the model on all tokens in a held-out 10\% of
the documents, and calculate F-2 score (preferring recall over precision) using
0.5 as the binarization threshold of model output.
We use an early stopping threshold of 1e-05 on this F-2 score, with a patience
of 5 epochs and a maximum of 50 epochs of training.

\subsection{Post-processing methods}
\label{sec:post-processing}

Given a set of token-level relevance annotations, HARE
provides three post-processing techniques for analyzing and improving
annotation results.

\textbf{Decision thresholding} The threshold for binarizing token
relevance scores is configurable between 0 and 1, to support more or less
conservative interpretation of model output; this is akin to exploring the
precision/recall curve. Figure~\ref{fig:thresholding} shows precision, recall,
and F-2 for different thresholding values from our mobility experiments,
using scores from ELMo embeddings.

\textbf{Collapsing adjacent segments} We consider any contiguous sequence
of tokens with scores at or above the binarization threshold to be a relevant
\textit{segment}. As shown in Figure~\ref{fig:collapsing}, multiple segments
may be interrupted by irrelevant tokens such as punctuation, or by noisy
relevance scores falling below the binarization threshold. As multiple adjacent segments may inflate a document's
overall relevance, our system includes a setting to collapse any adjacent
segments that are separated by $k$ or fewer tokens into a single segment.

\input{figures/fig-collapsing}

\input{figures/fig-viterbi-smoothing}

\textbf{Viterbi smoothing} By modeling token-level decisions as conditionally
independent of one another given the input features, we avoid assumptions
of strict segment bounds, but introduce some noisy output, as shown in
Figure~\ref{fig:viterbi-smoothing}. To reduce some of this noise, we
include an optional smoothing component based on the Viterbi algorithm.

\input{figures/fig-annotation-viewer}

We model the ``relevant''/``irrelevant'' state sequence discriminatively, using
annotation model outputs as state probabilities for each timestep, and calculate
the binary transition probability matrix by counting transitions in the training
data. We use these estimates to decode the most likely relevance state sequence
$R$ for a tokenized line $T$ in an input document, along with the corresponding path
probability matrix $W$, where $W_{j,i}$ denotes the likelihood of being in state
$j$ at time $i$ given $r_{i-1}$ and $t_i$. In order to produce continuous scores for
each token, we then backtrace through $R$ and assign score
$s_i$ to token $t_i$ as the conditional probability that $r_i$ is ``relevant'', given
$r_{i-1}$. Let $Q_{j,i}$ be the likelihood of transitioning from state $R_{i-1}$ to
$j$, conditioned on $T_i$, as:
\vspace{-0.2cm}
\begin{equation}
    Q_{j,i} = \frac{W_{j,i}}{W_{R_{i-1},i-1}}
\vspace{-0.1cm}
\end{equation}
The final conditional probability $s_i$ is calculated by normalizing over
possible states at time $i$:
\vspace{-0.2cm}
\begin{equation}
    s_i = \frac{Q_{1,i}}{Q_{0,i}+Q_{1,i}}
\vspace{-0.1cm}
\end{equation}
These smoothed scores can then be binarized using the configurable decision
threshold.

\subsection{Annotation viewer}
\label{ssec:annotation-viewer}

Annotations on any individual document can be viewed using a web-based
interface, shown in Figure~\ref{fig:annotation-viewer}. All tokens with scores
at or above the decision threshold are highlighted in yellow, with each
contiguous segment shown in a single highlight. Configuration settings for
post-processing methods are
provided, and update the displayed annotations when changed. On click,
each token will display the score assigned to it by the annotation model after
post-processing.
If the document being viewed is labeled with gold annotations, these are shown
in bold red text. Additionally, document-level summary statistics and
evaluation measures, with current post-processing, are
displayed next to the annotations.

\input{figures/fig-ranking-interface}

\subsection{Document set ranking}
\label{ssec:document-ranking}

\subsubsection{Ranking methods}

Relevance scoring methods are highly task-dependent, and may
reflect different priorities such as information density or diversity of
information returned.
In this system, we
provide three general-purpose relevance scorers, each of which operates after
any post-processing.

\textbf{Segments+Tokens} Documents are scored by multiplying their number of
relevant segments by a large constant and adding the number of relevant tokens
to break any ties by segment count. As relevant information may be sparse,
no normalization by document length is used.

\textbf{SumScores} Documents are scored by summing the continuous relevance
scores assigned to all of their tokens. As with the Segments+Tokens scorer,
no adjustment is made for document length.

\textbf{Density} Document scores are the ratio of binarized relevant tokens
to total number of tokens.

The same scorer can be used to rank gold annotations and
model annotations, or different scorers can be chosen.  Ranking quality is
evaluated using Spearman's $\rho$, which ranges
from -1 (exact opposite ranking) to +1 (same ranking), with 0 indicating no
correlation between rankings. We use Segments+Tokens as default; a comparison
of ranking methods is in Appendix~\ref{app:ranking}.

\subsubsection{Ranking interface}

Our system also includes a web-based ranking interface, which displays the
scores and corresponding ranking assigned to a set of annotated documents, as
shown in Figure~\ref{fig:ranking-interface}. For ease of visual distinction, we
include colorization of rows based on configurable score thresholds. Ranking
methods used for model scores and gold annotations (when present) can be
adjusted independently, and our post-processing methods 
(Section~\ref{sec:post-processing}) can also be adjusted to affect ranking.

\subsection{Qualitative analysis tools}
\label{ssec:qualitative-analysis}

We provide a set of three tools for performing qualitative analysis of
annotation outcomes. The first measures lexicalization of each unique token
in the dataset with respect to relevance score, by averaging the assigned
relevance score (with or without smoothing) for each instance of each token. Tokens
with a frequency below a configurable minimum threshold are excluded.

Our other tools analyze the aggregate relevance score patterns in an annotation
set. For labeled data, as shown in Figure~\ref{fig:thresholding}, we provide a
visualization of precision, recall, and F-2 when varying the
binarization threshold, including identifying the optimal threshold with respect
to F-2. We also include a label-agnostic analysis of patterns in output
relevance scores, illustrated in Figure~\ref{fig:score-patterns}, as a way to
evaluate the confidence of the annotator. Both of these tools are provided at
the level of an annotation set and individual documents.

\subsection{Implementation details}

Our automated annotation, post-processing, and document ranking algorithms
are implemented in Python, using the NumPy and Tensorflow libraries. Our
demonstration interface is implemented using the Flask library, with all
backend logic handled separately in order to support modularity of the user
interface.

%% file: tables/tbl-btris.tex
\begin{table}[t]
    \centering
    {\small
    \begin{tabular}{r|cc}
        &SpaCy&WordPiece\\
        \hline
        Num documents&\multicolumn{2}{c}{400}\\
        Avg tokens per doc&537&655\\
        Avg mobility tokens per doc&97&112\\
        Avg mobility segments per doc&\multicolumn{2}{c}{9.2}\\
    \end{tabular}
    }
    \caption{Statistics for dataset of mobility information, using SpaCy and
             WordPiece tokenization.}
    \label{tbl:btris}
\end{table}

%% file: figures/fig-workflow.tex
\begin{figure}[t]
    \centering
    \includegraphics[width=0.45\textwidth]{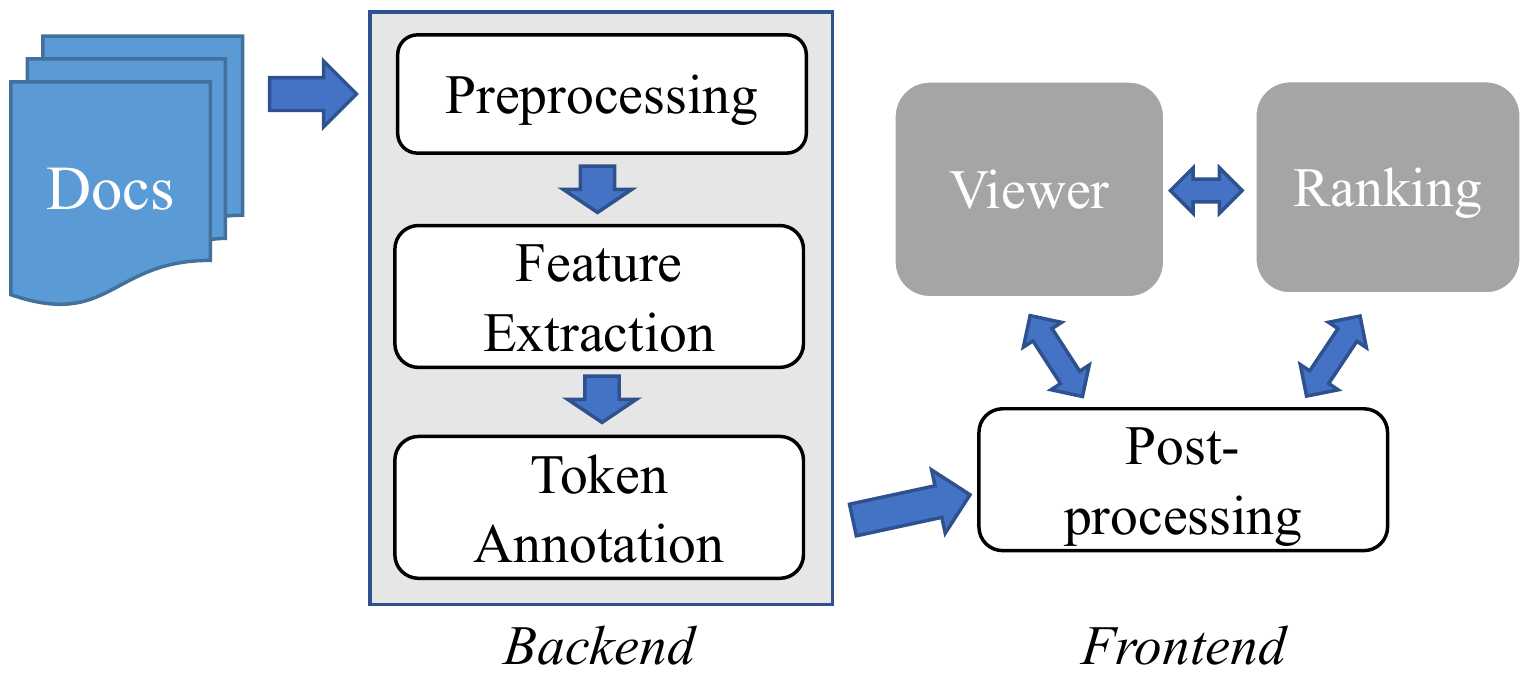}
    \caption{HARE workflow for working with a set of documents;
             outlined boxes indicate automated components,
             and gray boxes signify user interfaces.}
    \label{fig:workflow}
\end{figure}

%% file: figures/fig-thresholding.tex
\begin{figure}[t]
    \centering
    \includegraphics[width=0.45\textwidth]{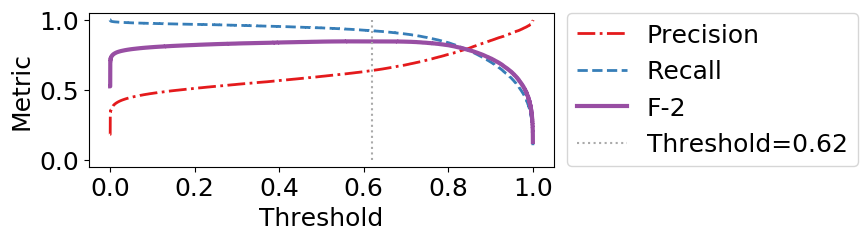}
    \caption{Precision, recall, and F-2 when varying binarization threshold
             from 0 to 1, using ELMo embeddings. The threshold corresponding to
             the best F-2 is marked with a dotted vertical line.}
    \label{fig:thresholding}
\end{figure}

%% file: figures/fig-collapsing.tex
\begin{figure}[t]
    \centering
    \begin{subfigure}[t]{0.45\textwidth}
        \centering
        \includegraphics[width=\textwidth]{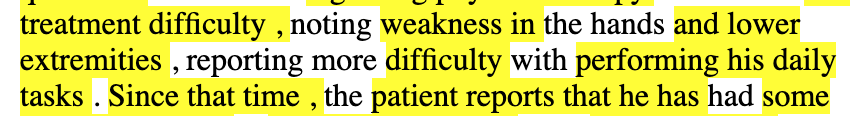}
        \caption{No collapsing}
    \end{subfigure}
    \begin{subfigure}[t]{0.45\textwidth}
        \centering
        \includegraphics[width=\textwidth]{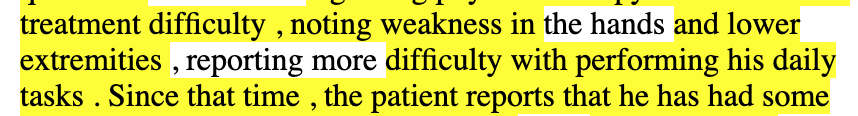}
        \caption{Collapse one blank}
    \end{subfigure}
    \caption{Collapsing adjacent segments illustration.}
    \label{fig:collapsing}
\end{figure}

%% file: figures/fig-viterbi-smoothing.tex
\begin{figure}[b]
    \centering
    \begin{subfigure}[t]{0.45\textwidth}
        \centering
        \includegraphics[width=\textwidth]{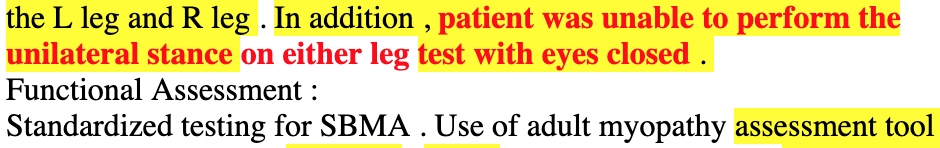}
        \caption{Without smoothing}
    \end{subfigure}
    \begin{subfigure}[t]{0.45\textwidth}
        \centering
        \includegraphics[width=\textwidth]{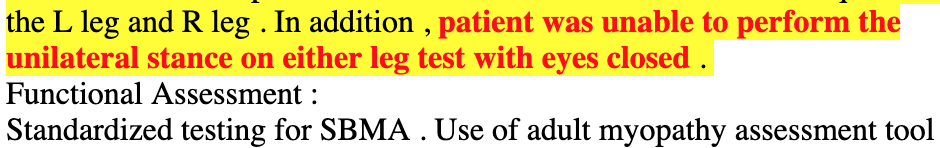}
        \caption{With smoothing}
    \end{subfigure}
    \caption{Illustration of Viterbi smoothing.}
    \label{fig:viterbi-smoothing}
\end{figure}

%% file: figures/fig-annotation-viewer.tex
\begin{figure}[t]
    \centering
    \includegraphics[width=0.45\textwidth]{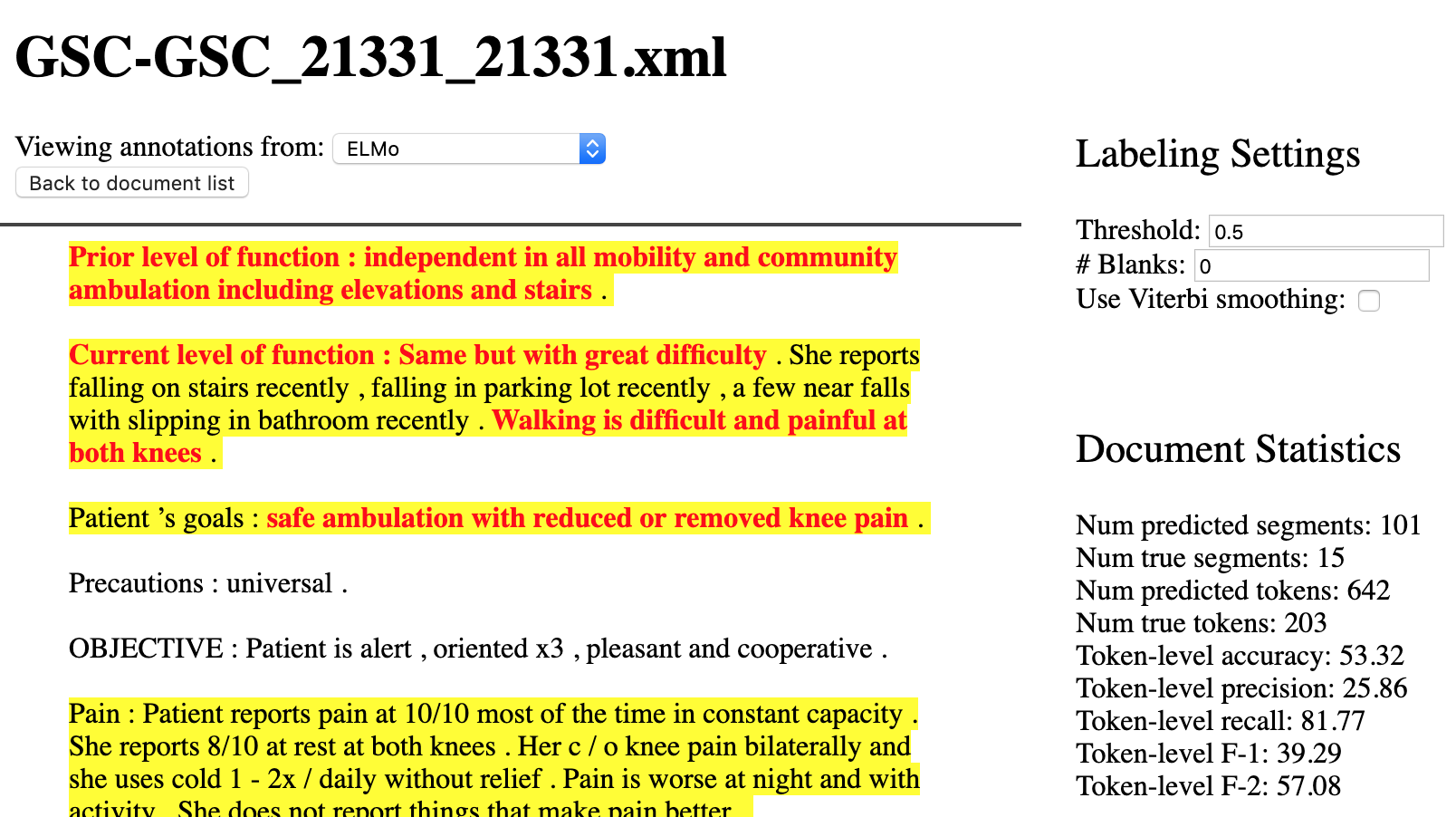}
    \caption{Annotation viewer interface.}
    \label{fig:annotation-viewer}
\end{figure}

%% file: figures/fig-ranking-interface.tex
\begin{figure}[t]
    \centering
    \includegraphics[width=0.45\textwidth]{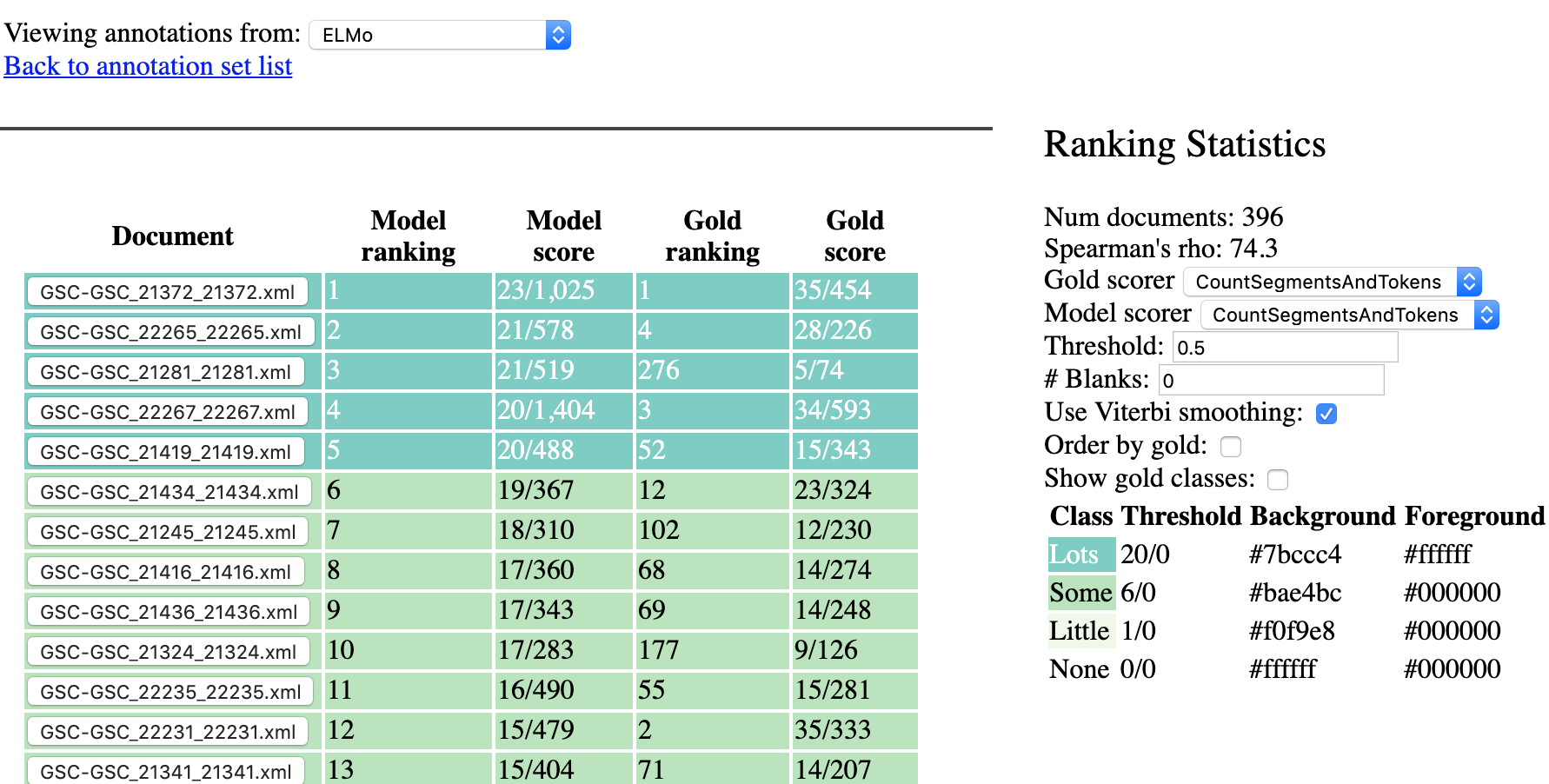}
    \caption{Ranking interface.}
    \label{fig:ranking-interface}
\end{figure}

%% file: sections/4-results.tex
\input{tables/tbl-results}

\section{Results on mobility}

Table~\ref{tbl:results} shows the token-level annotation and document ranking
results for our experiments on mobility information. Static and
contextualized embedding models performed equivalently well on token-level annotations;
BERT embeddings actually underperformed static embeddings and ELMo on both
precision and recall. Interestingly, static embeddings yielded the best ranking
performance of $\rho=0.862$, compared to $0.771$ with ELMo and $0.689$ with BERT.
Viterbi smoothing makes a minimal difference in token-level tagging, but increases
ranking performance considerably, particularly for contextualized models. It also
produces a qualitative improvement by trimming out extraneous tokens at the start
of several segments, as reflected by the improvements in precision.

The distribution of token scores from each model (Figure~\ref{fig:score-patterns})
shows that all three embedding models yielded a
roughly bimodal distribution, with most scores in the ranges $[0,0.2]$ or $[0.7,1.0]$.

\input{figures/fig-score-patterns}

%% file: tables/tbl-results.tex
\begin{table}[t]
    \centering
    \setlength{\tabcolsep}{3.0pt}
    {\small
    \begin{tabular}{cc|ccc|c}
        \multirow{2}{*}{Embeddings}&\multirow{2}{*}{Smoothing}&\multicolumn{3}{c}{Annotation}&\multicolumn{1}{c}{Ranking}\\
                                 &   &Pr&Rec&F-2&$\rho$\\
        \hline
        \multirow{2}{*}{Static}  &No &59.0&94.7&84.4&0.862\\
                                 &Yes&60.5&93.7&84.3&0.899\\
        \hline
        \multirow{2}{*}{ELMo}    &No &60.2&94.1&84.4&0.771\\
                                 &Yes&66.5&91.4&84.8&0.886\\
        \hline
        \multirow{2}{*}{BERT}    &No &55.3&93.8&82.2&0.689\\
                                 &Yes&62.3&90.8&84.3&0.844\\
    \end{tabular}
    }
    \caption{Annotation and ranking evaluation results on mobility documents,
             using three embedding sources. Results are given with and without
             Viterbi smoothing, using binarization threshold=0.5 and no
             collapsing of adjacent segments. Pr=precision, Rec=recall,
             $\rho$=Spearman's $\rho$ Pr/Rec/F2 are macro-averaged over folds,
             $\rho$ is over all test predictions.}
    \label{tbl:results}
\end{table}

%% file: figures/fig-score-patterns.tex
\begin{figure}[t]
    \centering
    \includegraphics[width=0.45\textwidth]{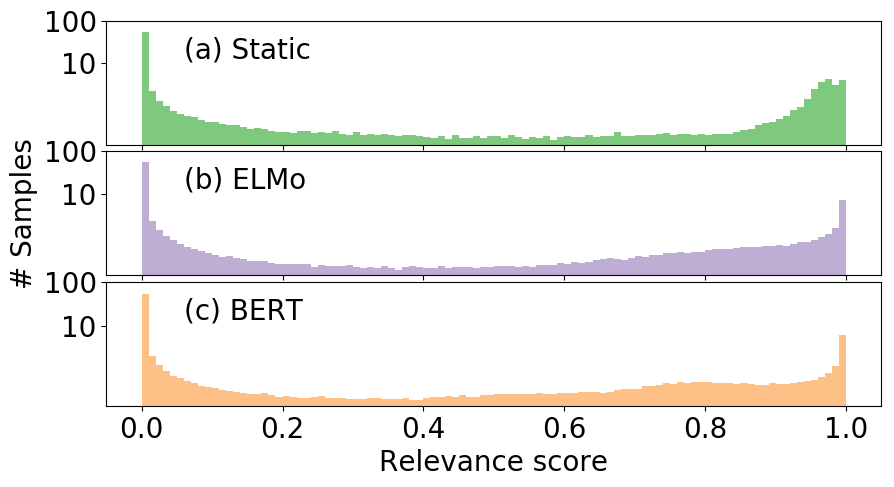}
    \caption{Distribution of token relevance scores on mobility data:
             (a) word2vec, (b) ELMo, and (c) BERT.}
    \label{fig:score-patterns}
\end{figure}

%% file: sections/5-discussion.tex
\section{Discussion}

Though our system is designed to address different needs
from other NLP annotation tools, components such as annotation viewing are
also addressed in other established systems. Our implementation decouples
backend analysis from the front-end interface; in future work, we plan to
add support for integrating our annotation and ranking systems into existing
platforms such as brat. Our tool can also easily be extended to both
multi-class and multilabel applications; for a detailed discussion, see
Appendix~\ref{app:extension}.

In terms of document ranking methods, it may be preferred to rank
documents jointly instead of independently, in order to account for challenges
such as duplication of information
(common in clinical data; \citet{Taggart2015}) or subtopics.
However, these decisions are highly task-specific, and are an important focus
for designing ranking utility within specific domains.

%% file: sections/6-conclusions.tex
\section{Conclusions}

We introduced HARE, a supervised system for highlighting relevant information
and interactive exploration of model outcomes. We demonstrated its utility in
experiments with clinical records annotated for narrative descriptions of
mobility status. We also provided qualitative analytic tools for
understanding the outcomes of different annotation models. In future work, we
plan to extend these analytic tools to provide rationales
for individual token-level decisions. Additionally, given the clear importance
of contextual information in token-level annotations, the static transition
probabilities used in our Viterbi smoothing technique are likely to degrade its
effect on the output. Adding support for dynamic, contextualized
estimations of transition probabilities will provide more fine-grained modeling
of relevance, as well as more powerful options for post-processing.

Our system is available online at \url{https://github.com/OSU-slatelab/HARE/}.
This research was supported by the Intramural Research Program of the
National Institutes of Health and the US Social Security Administration.

%% file: sections/appendix.tex
\section{Hyperparameters}
\label{app:hyperparameters}

This section describes each of the settings evaluated for the various
hyperparameters used in our experiments on mobility information. 
We first experimented with different pretrained embeddings used for each of
our embedding model options; results are shown in Figure~\ref{fig:app-embedding-models}.

\textbf{Static embedding model} (Figure~\ref{fig:hp-static-models}) We
evaluated three commonly used benchmark embedding sets: word2vec skipgram \cite{Mikolov2013a}
using GoogleNews,\footnote{
    \url{http://google.com/archive/p/word2vec/}
}
FastText skipgram with subword information on WikiNews,\footnote{
    \url{https://fasttext.cc/docs/en/}
}
and GloVe \cite{Pennington2014} on 840 billion tokens of Common Crawl.\footnote{
    \url{http://nlp.stanford.edu/projects/glove/}
}
Additionally, we experimented with two in-domain embedding sets, trained on
10 years of Physical Therapy and Occupational Therapy records from the NIH
Clinical Center (referred to as ``PT/OT''), using word2vec skipgram and
FastText skipgram. word2vec GoogleNews embeddings produced the best dev F-2.

\textbf{ELMo model} (Figure~\ref{fig:hp-elmo-models}) We experimented with
three pretrained ELMo models:\footnote{
    All downloaded from \url{https://allennlp.org/elmo}
} the ``Original'' model trained on the 1 Billion Word Benchmark, the
``Original (5.5B)'' model trained with the same settings on Wikipedia and
machine translation data, and a model trained on PubMed abstracts. The
Original (5.5B) model produced the best dev F-2.

\textbf{BERT model} (Figure~\ref{fig:hp-bert-models}) We experimented with
three pretrained BERT models: BERT-Base,\footnote{
    \url{https://github.com/google-research/bert}
} BioBERT \cite{Lee2019} (v1.1) trained on 1 million PubMed abstracts,\footnote{
    \url{https://github.com/naver/biobert-pretrained}
} and clinicalBERT \cite{Alsentzer2019} trained on MIMIC data.\footnote{
    \url{https://github.com/EmilyAlsentzer/clinicalBERT}
}
We use uncased versions of BERT-Base and clinicalBERT, as casing is not a
reliable signal in clinical data; BioBERT is only available in a cased version.
clinicalBERT produced the best dev F-2.

Once the best embedding models for each method were identified, we experimented
with network and training hyperparameters, with results shown in
Figure~\ref{fig:app-hyperparameters}.

\textbf{Irrelevant:relevant sampling ratio} (Figure~\ref{fig:hp-neg-ratio})
We experimented with the ratio of irrelevant to relevant samples drawn for
each training epoch in ${0.25, 0.5, 0.75, 1, 1.5, 2, 2.5, 3}$. A ratio of
0.75 gave the best dev F-2.

\textbf{Positive fraction} (Figure~\ref{fig:hp-pos-fraction}) We varied the
fraction of total dataset positive samples drawn for each training epoch
from 10\% to 100\% at intervals of 10\%. The best dev F-2 was produced by
using all positive samples in each epoch.

\textbf{Dropout rate} (Figure~\ref{fig:hp-dropout}) We experimented with
an input dropout rate from 0\% to 90\%, at intervals of 10\%; the best 
results were produced with a 60\% dropout rate.

\textbf{Weighting scheme} (Figure~\ref{fig:hp-class-weights}) Given the imbalance
of relevant to irrelevant samples in our dataset, we experimented with
weighting relevant samples by a factor of 1 (equal weight), 2, 3, 4, and 5.
A weighting of 2:1 produced the best dev F-2.

\textbf{Hidden layer configuration} (Figure~\ref{fig:hp-layers}) We
experimented with the configuration of our DNN model, using hidden
layer size $\in \{10, 100, 300\}$ and number of layers from 1 to 3.
The best dev F-2 results were achieved using 3 hidden layers of size
300.

\section{Ranking methods}
\label{app:ranking}

A comparison of ranking methods used for model and gold scores is provided
in Figure~\ref{fig:app-ranking-methods}. We found that for our experiments,
Segments+Tokens and SumScores correlated fairly well with one another, but Density,
due to its normalization for document length, works best when used to rank
both model and gold scores. SumScores provided the best overall ranking
correlation; however, we use Segments+Tokens as the default setting for our
system for its clear interpretation.

\section{Extending to multi-class/multilabel applications}
\label{app:extension}

Our experiments focused on binary relevance with respect to mobility information.
However, our system can be fairly straightfowardly extended to both multi-label
(i.e., multiple relevance criteria) and multi-class (e.g., NER) settings.

For multi-label settings, such as looking for evidence of limitations in either
mobility or interpersonal interactions, the only requirement is having data that
are annotated for each relevance criterion. These can be the same data with
multiple annotations, or different datasets; in either case, binary relevance
annotators can be trained independently for each specific relevance criterion.
Our post-processing components such as Viterbi smoothing can then be applied
independently to each set of relevance annotations as desired. The primary
extension required would be to the visualization interface, to support display
of multiple (potentially overlapping) annotations. Alternatively, our modular
handling of relevance annotations could be redirected to another visualization
interface with existing support for multiple annotations, such as brat.

Extending to multi-class settings would require fairly minimal updates to both
the interface and our relevance annotation model. Our model is trained using
two-class cross (relevant and irrelevant) cross-entropy; this could easily be
extended to $n$-ary cross entropy for any desired number of classes, and trained
with token-level data annotated with the appropriate classes. In terms of
visualization and analysis, the two modifications required would be adding
differentiating displays for the different classes annotated (e.g., different
colors), and updating the displayed evaluation statistics to micro/macro
evaluations over the multiple classes. Qualitative analysis features such as
relevance score distribution and lexicalization are already dependent only on
the scores assigned to the ``relevant'' class, and could be presented for each
class independently.

\input{figures/fig-app-embedding-models}
\input{figures/fig-app-hyperparameters}
\input{figures/fig-app-ranking-methods}

%% file: figures/fig-app-embedding-models.tex
\begin{figure*}[t]
    \centering
    \begin{subfigure}[t]{0.32\textwidth}
        \includegraphics[width=\textwidth]{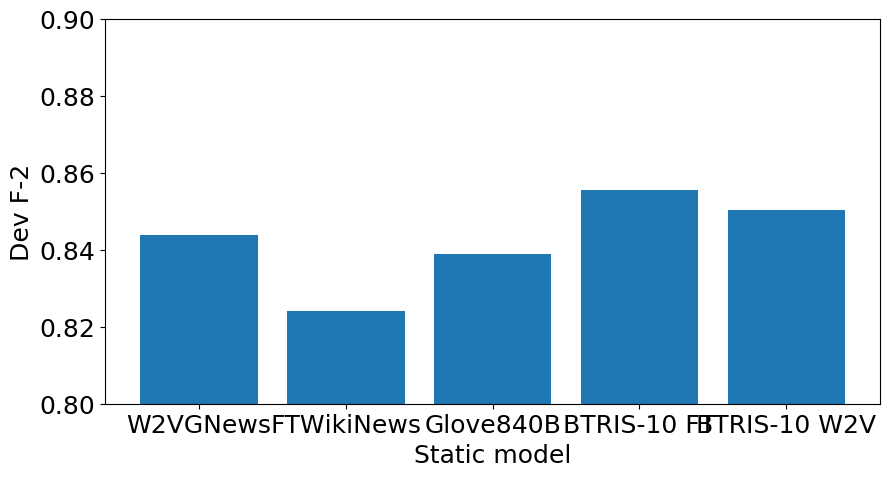}
        \caption{Static word embedding models}
        \label{fig:hp-static-models}
    \end{subfigure}
    \begin{subfigure}[t]{0.32\textwidth}
        \includegraphics[width=\textwidth]{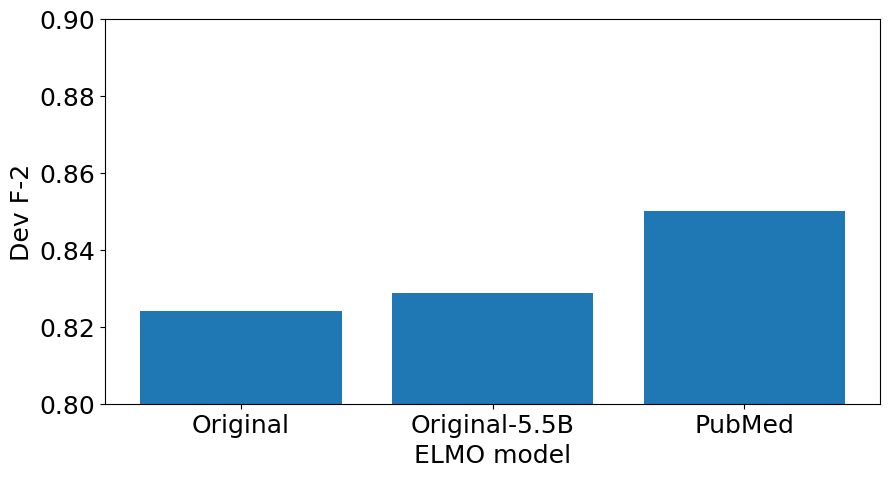}
        \caption{ELMo embedding models}
        \label{fig:hp-elmo-models}
    \end{subfigure}
    \begin{subfigure}[t]{0.32\textwidth}
        \includegraphics[width=\textwidth]{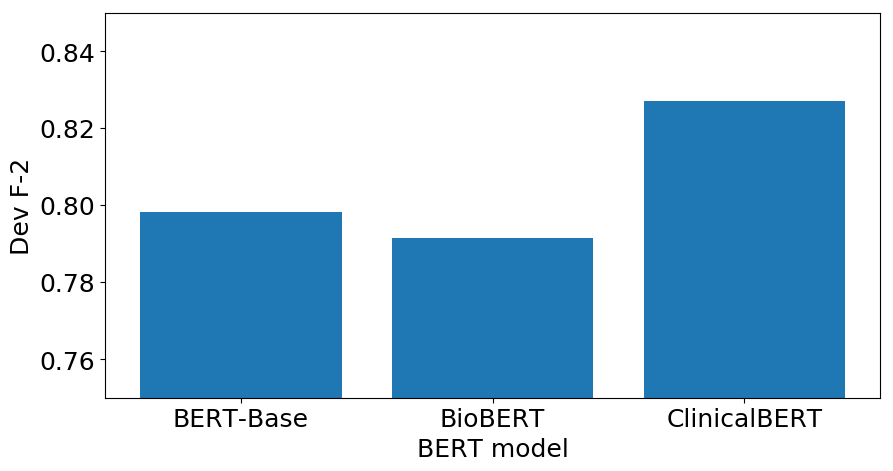}
        \caption{BERT embedding models}
        \label{fig:hp-bert-models}
    \end{subfigure}
    \caption{Embedding model selection results, by F-2 on cross validation
             development set.
             Default settings for other hyperparameters were: relevant:irrelevant ratio of 1:1,
             sampling 50\% of positive samples
             per epoch, dropout of 0.5, equal class weights, and DNN configuration
             one 100-dimensional hidden layer.}
    \label{fig:app-embedding-models}
\end{figure*}

%% file: figures/fig-app-hyperparameters.tex
\begin{figure*}[t]
    \centering
    \begin{subfigure}[t]{0.45\textwidth}
        \includegraphics[width=\textwidth]{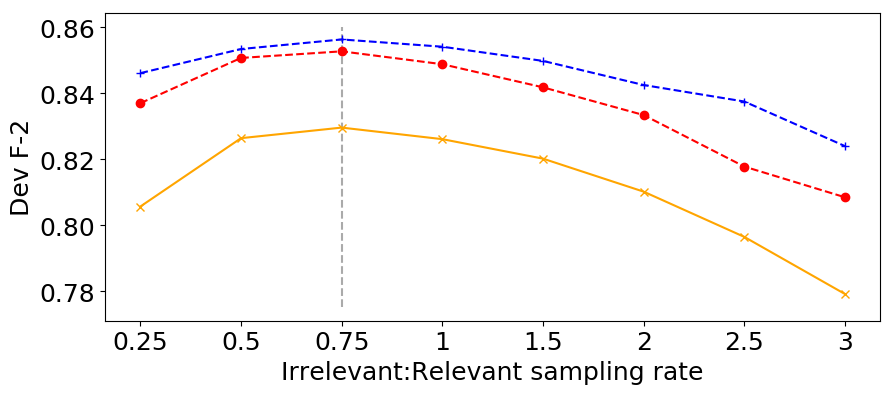}
        \caption{Irrelevant:relevant sampling ratio in training}
        \label{fig:hp-neg-ratio}
    \end{subfigure}
    \begin{subfigure}[t]{0.45\textwidth}
        \includegraphics[width=\textwidth]{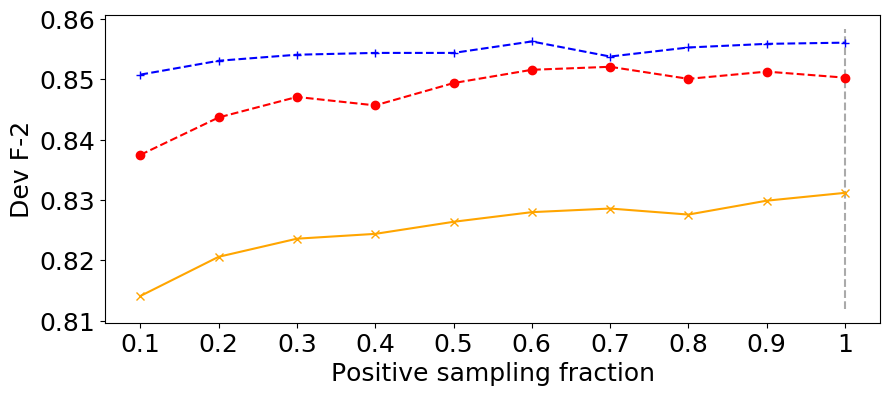}
        \caption{Fraction of relevant samples per epoch}
        \label{fig:hp-pos-fraction}
    \end{subfigure}
    \begin{subfigure}[t]{0.45\textwidth}
        \includegraphics[width=\textwidth]{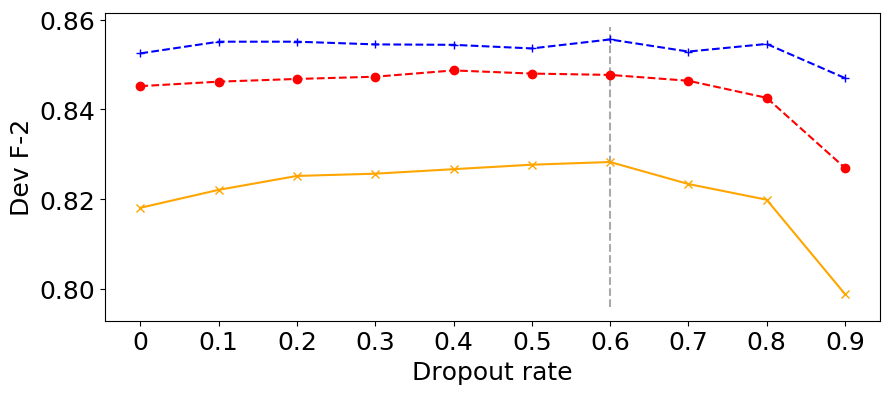}
        \caption{Dropout}
        \label{fig:hp-dropout}
    \end{subfigure}
    \begin{subfigure}[t]{0.45\textwidth}
        \includegraphics[width=\textwidth]{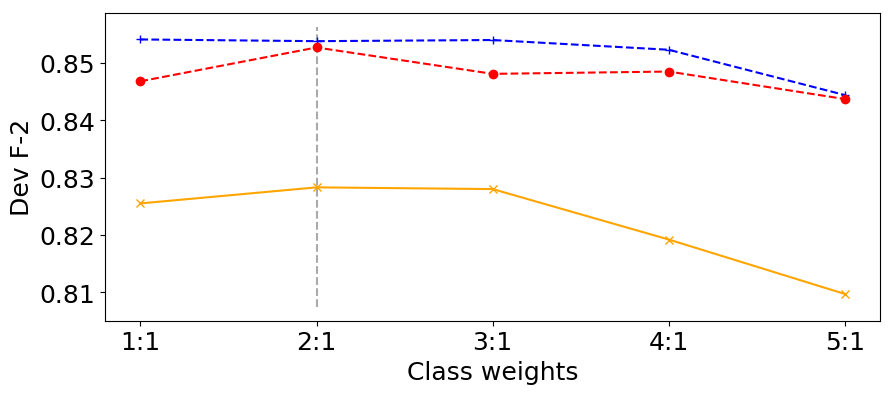}
        \caption{Weighting scheme (Relevant:Irrelevant)}
        \label{fig:hp-class-weights}
    \end{subfigure}
    \begin{subfigure}[t]{0.45\textwidth}
        \includegraphics[width=\textwidth]{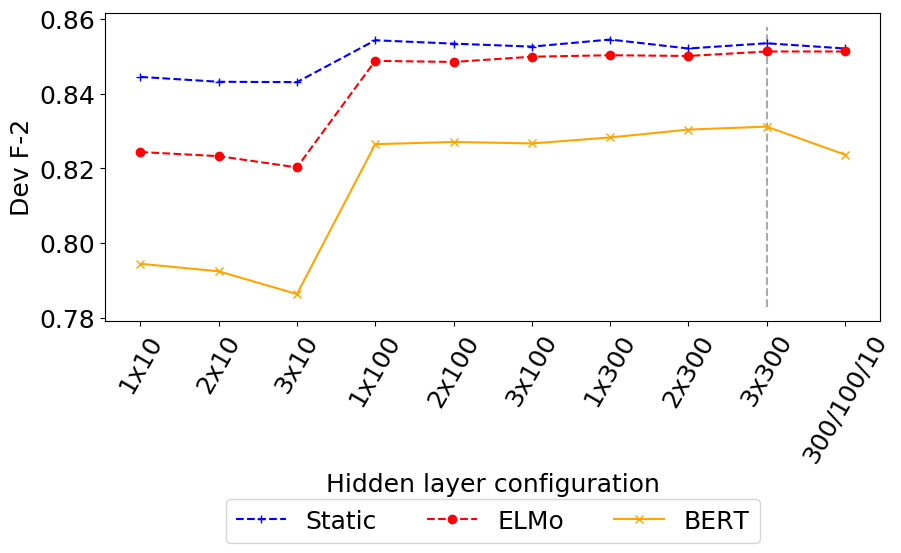}
        \caption{Configurations of DNN token annotator; $NxM$ indicates $N$
                 layers of dimensionality $M$; $M_1/M_2$ indicates
                 that the first hidden layer is of dimensionality $M_1$, the
                 second of $M_2$.}
        \label{fig:hp-layers}
    \end{subfigure}
    \caption{Hyperparameter tuning results, measuring F-2 on development set in
             cross-validation experiments. For each embedding method, the best
             model (shown in Figure~\ref{fig:app-embedding-models}) was used.
             All other hyperparameter setting defaults were as
             described in Figure~\ref{fig:app-embedding-models}.
             The best-performing setting 
             for each hyperparameter, determined by the mean Dev F-2 across all three
             embedding methods, is indicated with a vertical dashed line.}
    \label{fig:app-hyperparameters}
\end{figure*}

%% file: figures/fig-app-ranking-methods.tex
\begin{figure*}[t]
    \centering
    \begin{subfigure}[t]{0.45\textwidth}
        \includegraphics[width=\textwidth]{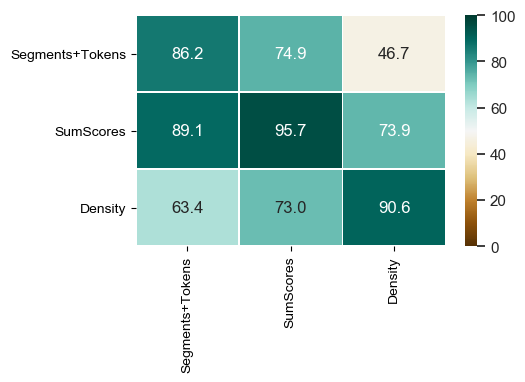}
        \caption{Static (no smoothing)}
        \label{fig:ranking-static-no-smoothing}
    \end{subfigure}
    \begin{subfigure}[t]{0.45\textwidth}
        \includegraphics[width=\textwidth]{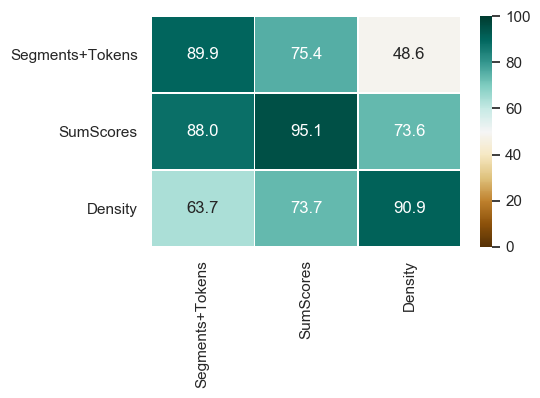}
        \caption{Static (with smoothing)}
        \label{fig:ranking-static-with-smoothing}
    \end{subfigure}
    \begin{subfigure}[t]{0.45\textwidth}
        \includegraphics[width=\textwidth]{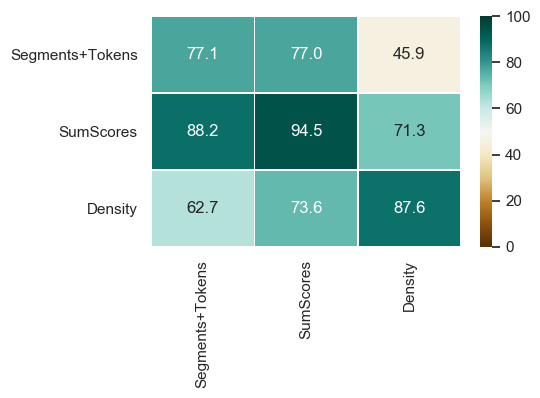}
        \caption{ELMo (no smoothing)}
        \label{fig:ranking-elmo-no-smoothing}
    \end{subfigure}
    \begin{subfigure}[t]{0.45\textwidth}
        \includegraphics[width=\textwidth]{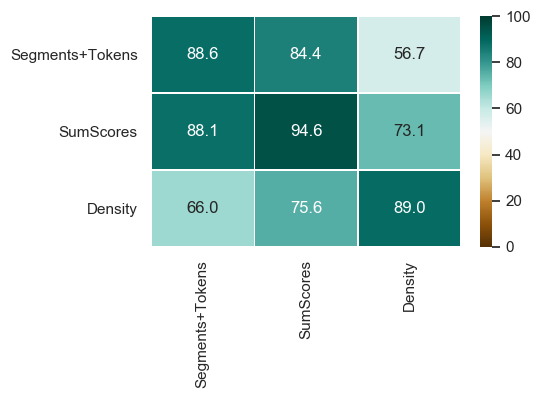}
        \caption{ELMo (with smoothing)}
        \label{fig:ranking-elmo-with-smoothing}
    \end{subfigure}
    \begin{subfigure}[t]{0.45\textwidth}
        \includegraphics[width=\textwidth]{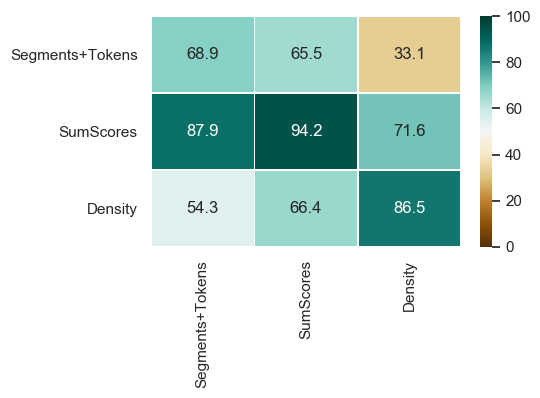}
        \caption{BERT (no smoothing)}
        \label{fig:ranking-bert-no-smoothing}
    \end{subfigure}
    \begin{subfigure}[t]{0.45\textwidth}
        \includegraphics[width=\textwidth]{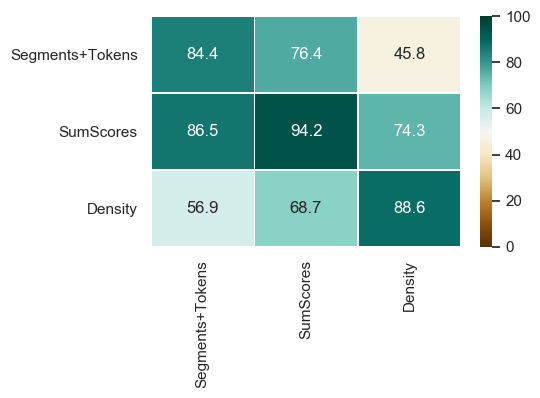}
        \caption{BERT (with smoothing)}
        \label{fig:ranking-bert-with-smoothing}
    \end{subfigure}
    \caption{Comparison of ranking methods used for model scores and gold
             scores. Scores given are Spearman's rank correlation coefficient
             ($\rho$). System outputs using all three embedding methods are
             compared, both with and without Viterbi smoothing.}
\end{figure*}